\pdfoutput=1
\documentclass[11pt]{article}

\usepackage{acl}
\usepackage{times}
\usepackage{latexsym}
\usepackage{hyperref}
\usepackage{framed, color}
\usepackage{graphicx}
\usepackage{booktabs}
\usepackage{todonotes}
\usepackage{multirow}
\usepackage{comment}
\usepackage{adjustbox}
\usepackage{caption}
\usepackage{subcaption}
\usepackage{tipa} 
\usepackage{enumitem}
\usepackage{amssymb} % for checkmark
\usepackage{pifont} % for xmark
\usepackage{array}
\usepackage{tabularx}
\usepackage{float}
\usepackage{verbatim}

\usepackage[T1]{fontenc}
\usepackage[utf8]{inputenc}

\usepackage{microtype}
\definecolor{light-gray}{gray}{0.95}
\newcommand{\code}[1]{\colorbox{light-gray}{\texttt{#1}}}

\definecolor{srclarge}{gray}{1.0}
\definecolor{srcsmall}{gray}{1.0}
\definecolor{large}{rgb}{0.94, 0.87, 0.8}
\definecolor{small}{rgb}{0.67, 0.9, 0.93}

\usepackage{xspace}
\newcommand*{\yoruba}{Yor\`ub\'a\xspace}

\title{MMTAfrica: Multilingual Machine Translation for African Languages}

\author{Chris C. Emezue\thanks{$\_$Authors contributed equally to this work. Correspondence to \textit{chris.emezue@gmail.com} or \textit{femipancrace.dossou@gmail.com}} \\
  Technical University of Munich \\
  Mila Quebec AI Institute\thanks{$\_$Independent Research done while interning at Mila Quebec AI Institute}\\
  Masakhane \\\And
  Bonaventure F. P. Dossou\footnotemark[1] \\
  Jacobs University Bremen\\
  Mila Quebec AI Institute\footnotemark[2] \\
  Masakhane \\
  }

\begin{document}
\maketitle
\begin{abstract}
%Multilingual Machine Translation (MMT) has made lots of progress in the recent years, especially for low-resource languages. African languages, over 2000, are truly complex and low-resourced, yet there has been considerably little effort on MMT on them. 
In this paper, we focus on the task of multilingual machine translation for African languages and describe our contribution in the 2021 WMT Shared Task: Large-Scale Multilingual Machine Translation. We introduce MMTAfrica, the first many-to-many multilingual translation system for six African languages: Fon (fon), Igbo (ibo), Kinyarwanda (kin), Swahili/Kiswahili (swa), Xhosa (xho), and Yoruba (yor) and two non-African languages: English (eng) and French (fra). For multilingual translation concerning African languages, we introduce a novel backtranslation and reconstruction objective, BT\&REC, inspired by the random online back translation and T5 modeling framework respectively, to effectively leverage monolingual data. Additionally, we report improvements from MMTAfrica over the FLORES 101 benchmarks (spBLEU gains ranging from $+0.58$ in Swahili to French to $+19.46$ in French to Xhosa). We release our dataset and code source at \url{https://github.com/edaiofficial/mmtafrica}.
\end{abstract}
% https://github.com/edaiofficial/mmtafrica

In this paper, we make use of the following notations:
 \begin{itemize}
     \item  $\circledast$ refers to any language in the set $\{eng,fra,ibo,fon,swa,kin,xho,yor\}$.
     \item  $\diamond$ refers to any language in the set $\{eng,fra,ibo,fon\}$.
     \item  AL(s) refers to African language(s).
     \item X$\longrightarrow$ Y refers to neural machine translation from language X to language Y.
\end{itemize}
% \section{Outline}

% \begin{itemize}
%     \item Related Works: 
%     \begin{itemize}
%         \item few works on MT african langs. Talk about African langs being understudied (with proof from \cite{joshi}
%         \item works on mmt in general....the mt5 model, mbart, etc.
%         \item works on leveraging BT and monolingual data in reconstruction.
%         \item end by saying our work combines everything into one and analyses each of them
%     \end{itemize}
% \end{itemize}

\section{Introduction}
Despite the progress of multilingual machine translation (MMT) and the many efforts towards improving its performance for low-resource languages, African languages suffer from under-representation. For example, of the $2000$ known African languages~\cite{ethnologue} only $17$ of them are available in the FLORES 101 Large-scale Multilingual Translation Task as at the time of this research. Furthermore, most research that look into transfer learning of multlilingual models from high-resource to low-resource languages rarely work with ALs in the low-resource scenario. While the consensus is that the outcome of the research made using the low-resource non-African languages should be scalable to African languages, this cross-lingual generalization is not guaranteed\cite{masakhanecommunity} and the extent to which it actually works remains largely under-studied. Transfer learning from African languages to African languages sharing the same language sub-class has been shown to give better translation quality than from high-resource Anglo-centric languages~\cite{DBLP:journals/corr/abs-2104-00366} calling for the need to investigate AL$\longleftrightarrow$AL multilingual translation.
\raggedbottom

 We take a step towards addressing the under-representation of African languages in MMT and improving experiments by participating in the 2021 WMT Shared Task: Large-Scale Multilingual Machine Translation and focusing solely on ALs$\longleftrightarrow$ALs. We focused on $6$ African languages and $2$ non-African languages (English and French). Table~\ref{tab:language} gives an overview of our focus African languages in terms of their language family, number of speakers and the regions in Africa where they are spoken~\cite{masakhaner}.  We chose these languages in an effort to create some language diversity: the 6 African languages span the most widely and least spoken languages in Africa. 
 
 \begin{comment}
 Additionally, they have some similar, as well as contrasting, characteristics which offer interesting insights for future work in ALs:

\begin{itemize}
\item Igbo, \yoruba and Fon use diacritics in their language structure while Kinyarwanda, Swahili and Xhosa do not. Various forms of code-mixing are prevalent in Igbo~\cite{okwugbe}.
    \item Fon was particularly chosen because there is only a minuscule amount of online (parallel or monolingual) corpora compared to the other $5$ languages. We wanted to investigate and provide valuable insights on improving translation quality of very low-resourced African languages.
    \item Kinyarwanda and Fon are the only African languages in our work not covered in the FLORES Large-Scale Multilingual Machine Translation Task and also not included in the pretraining of the original model framework used for MMTAfrica. Based on this, we were able to understand the performance of multilingual translation finetuning involving languages not used in the original pretraining objective. We also offered a method to improve the translation quality of such languages.
\end{itemize}
\end{comment}
Our main contributions are summarized below:
\begin{enumerate}
\item MMTAfrica -- a many-to-many AL$\longleftrightarrow$AL multilingual model for $6$ African languages.
\item Our novel reconstruction objective (described in section~\ref{trainingobj}) and the \code{BT\&REC} finetuning setting, together with our proposals in section~\ref{baselineresults} offer a comprehensive strategy for effectively exploiting monolingual data of African languages in AL$\longleftrightarrow$AL multilingual machine translation, 
\item Evaluation of MMTAfrica on the \code{FLORES Test Set} reports significant gains in spBLEU over the  M2M MMT~\cite{m2m} benchmark model provided by \citet{flores101},
\item We further created a unique highly representative test set -- \code{MMTAfrica Test Set} -- and reported benchmark results and insights using MMTAfrica.
\end{enumerate}

\begin{table}[!h]
 \begin{center}
 \resizebox{\columnwidth}{!}{
 %\scalebox{0.90}{
 \footnotesize
  \begin{tabular}{p{14mm}p{20mm}p{33mm}p{9mm}rp{9mm}}
    \toprule
    % \multirow{2}{*}{\textbf{ISO}} &  \multirow{2}{*}{\textbf{Family}} & \multirow{2}{*}{\textbf{Speakers}} & \textbf{Region in}\\
    % & & & \textbf{Africa}\\
    \textbf{Language} &\textbf{Lang ID \newline (ISO 639-3)}&  \textbf{Family} & \textbf{Speakers} & \textbf{Region}\\
    \midrule
    Igbo &ibo& Niger-Congo-Volta-Niger & 27M & West \\
    \addlinespace[0.5em]
     Fon (Fongbe) &fon& Niger-Congo-Volta-Congo-Gbe & 1.7M & West \\
    \addlinespace[0.5em]
    Kinyarwanda & kin  & Niger-Congo-Bantu & 12M & East\\
   \addlinespace[0.5em]
    Swahili & swa & Niger-Congo-Bantu & 98M & Southern, Central \& East \\
   \addlinespace[0.5em]
    Xhosa &xho& Niger-Congo-Nguni Bantu & 19.2M & Southern \\
    \addlinespace[0.5em]
    \yoruba &yor&  Niger-Congo-Volta-Niger & 42M & West\\
    \bottomrule
  \end{tabular}
  }
  \caption{Language, family, number of speakers~\cite{ethnologue}, and regions in Africa. Adapted from \cite{masakhaner}}
  % Language codes are ISO 639-3.
  \label{tab:language}
  \end{center}
\end{table}

%Thanks to recent advances in deep learning (Sutskever et al., 2014; Bahdanau et al., 2015) and the availability of large-scale parallel corpora, machine translation has now reached impressive performance
%on several language pairs (Wu et al., 2016). However, these models work very well only
%when provided with massive amounts of parallel data, in the order of millions of parallel sentences.
%Unfortunately, parallel corpora are costly to build as they require specialized expertise, and are often
%nonexistent for low-resource languages. Conversely, monolingual data is much easier to find, and
%many languages with limited parallel data still possess significant amounts of monolingual data.

%Another pitfall of massively multilingual NMT is its poor zero-shot performance, particularly compared to pivot-based models. Without access to parallel training data for zero-shot language pairs, multilingual models easily fall into the trap of offtarget translation where a model ignores the given target information and translates into a wrong language

\section{Related Work}
\subsection{Multilingual Machine Translation (MMT)}
%When it comes to AL $\longrightarrow$ AL translation, one straight-forward approach here no or little parallel data is available is to use explicit bridging~\cite{johnson-etal-2017-googles}: translate to an intermediate language first (usually English) and then to the desired African target language. Two potential disadvantages of this approach are: a) the total translation time doubles, b) the potential loss of quality by translating to/from the intermediate language.

%The great success of the encoder-decoder~\cite{10.5555/2969033.2969173,DBLP:conf/emnlp/ChoMGBBSB14} NMT on bilingual datasets~\cite{bahd,transformer,wmt19,wmt20} inspired the extension of the original bilingual framework to handle more languages pairs simultaneously -- leading to multilingual neural machine translation. 

%Works on multilingual NMT have progressed from sharing the encoder for one-to-many translation~\cite{dong}, many-to-one translation~\cite{lee}, sharing the attention mechanism across multiple language pairs~\cite{multiway,dong} to optimizing a single NMT model (with a universal encoder and decoder) for the translation of multiple language pairs~\cite{DBLP:journals/corr/HaNW16,johnson-etal-2017-googles}. The universal encoder-decoder approach constructs a shared vocabulary for all languages in the training set, and uses just one encoder and decoder for multilingual translation between language pairs. \citet{johnson-etal-2017-googles} proposed to use a single model and prepend special symbols to the source text to indicate the target language. We adopt their model approach in this paper.

The current state of multilingual NMT, where a single NMT model is optimized for the translation of multiple language pairs~\cite{multiway,johnson-etal-2017-googles,lu-etal-2018-neural,aharoni-etal-2019-massively,mmtinthewild}, has become very appealing for a number of reasons. It is scalable and easy to deploy or maintan (the ability of a single model to effectively handle all translation directions from $N$ languages, if properly trained and designed, surpasses the scalability of $O(N^{2})$ individually trained models using the traditional bilingual framework). Multilingual NMT can encourage knowledge transfer among related language pairs~\cite{lakew-etal-2018-comparison,tan-etal-2019-multilingual} as well as positive transfer from higher-resource languages~\cite{zoph-etal-2016-transfer,neubig-hu-2018-rapid,DBLP:journals/corr/abs-1903-07091,aharoni-etal-2019-massively,johnson-etal-2017-googles} due to its shared representation, improve low-resource translation~\cite{DBLP:journals/corr/HaNW16,johnson-etal-2017-googles,mmtinthewild,mt5} and enable zero-shot translation (i.e. direct translation between a language pair never seen during training)~\cite{firat-etal-2016-zero,johnson-etal-2017-googles}.

Despite the many advantages of multilingual NMT it suffers from certain disadvantages. Firstly, the output vocabulary size is typically fixed regardless of the number of languages in the corpus and increasing the vocabulary size is costly in terms of computational resources because the training and inference time scales linearly with the size of the decoder’s output layer. 
%For example, the training dataset for all the languages in our work gave a total vocabulary size of $1,683,884$ tokens ($1,519,918$ with every sentence lowercased) but we were constrained to a decoder vocabulary size of $250,000$. 

Another pitfall of massively multilingual NMT is its poor zero-shot performance~\cite{firat-etal-2016-zero,DBLP:journals/corr/abs-1903-07091,johnson-etal-2017-googles,aharoni-etal-2019-massively}, particularly compared to pivot-based models (two bilingual models that translate from source to target language through an intermediate language), the spurious correlation issue~\cite{gu} and \textit{off-target translation}~\cite{johnson-etal-2017-googles} where the model ignores the given target information and translates into a wrong language. 

Our work is inspired by some research to improve the performance (including zero-shot translation) of multilingual models via back-translation and leveraging monolingual data. %have relied on leveraging the plentiful source and target side monolingual data that are available. For example, generating artificial parallel data with various forms of backtranslation~\cite{DBLP:journals/corr/SennrichHB15a} has been shown to greatly improve the overall (and zero-shot) performance of multilingual models~\cite{firat-etal-2016-zero,gu,lakew-etal-2018-comparison,DBLP:journals/corr/abs-2004-11867} as well as bilingual models~\cite{understandingbt}.
\citet{DBLP:journals/corr/abs-2004-11867} proposed random online backtranslation to enhance multilingual translation of unseen training language pairs. %Additionally, leveraging monolingual data by jointly learning to reconstruct the input while translating has been shown to improve neural machine translation quality~\cite{DBLP:journals/corr/abs-1809-02669,lampe,cheng-etal-2016-semi,zhang-zong-2016-exploiting}. 
\citet{DBLP:journals/corr/abs-2005-04816} leveraged monolingual data in a semi-supervised fashion and reported three major results:
\begin{enumerate}
    \item Using monolingual data significantly boosts the translation quality of low resource languages in multilingual models.
    \item Self-supervision improves zero-shot translation quality in multilingual models.
    \item Leveraging monolingual data with self-supervision provides a viable path towards adding new languages to multilingual models.
\end{enumerate}

%In this work, we build on all this by introducing an approach  building a baseline experiment (detailed in section \ref{baseline}) to test the effect of backtranslation on many-to-many translation between African languages.
%\subsection{MMT for African Languages}There is no surprise that when it comes to MMT, many (most of) approaches are English-centric. Therefore, when it comes to MMT for ALs, very few works have been done. Yet, among those few attempts, most focus on European Language$\rightarrow$AL, leaving behind the exploration and interesting approach of ALs-to-ALs. This paper is the first focusing solely on ALs-to-ALs, with the goal to draw useful insights and interest for future ALs-to-ALs translations research/work.
\section{Data Methodology}
Table \ref{tab:languagesize} presents the size of the gathered and cleaned parallel sentences for each language direction.
\begin{table*}[ht]
 \begin{center}
 \resizebox{0.7\textwidth}{!}{%
 %\scalebox{0.90}{
 \footnotesize
  \begin{tabular}{cccccccccc}
    \toprule
    % \multirow{2}{*}{\textbf{ISO}} &  \multirow{2}{*}{\textbf{Family}} & \multirow{2}{*}{\textbf{Speakers}} & \textbf{Region in}\\
    % & & & \textbf{Africa}\\
    &&&  &  &  \textbf{Target Language}&&& \\
    \addlinespace[0.5em]
  &\textbf{ibo}&  \textbf{fon}&  \textbf{kin}&  \textbf{xho}&  \textbf{yor}&  \textbf{swa}& \textbf{eng}&  \textbf{fra}\\
   \addlinespace[0.5em]
    \midrule
   \textbf{ibo} &- &{\setlength{\fboxsep}{0pt}\colorbox{srcsmall}{$3,179$}}& $52,685$ &$58,802$ & {\setlength{\fboxsep}{0pt}\colorbox{srclarge}{$134,219$}}&$67,785$&$85,358$&$57,458$ \\
    \addlinespace[0.5em]

    \textbf{fon} &$3,148$&-  &$3,060$ &$3,364$&$5,440$&$3,434$& {\setlength{\fboxsep}{0pt}\colorbox{srclarge}{$5,575$}}& {\setlength{\fboxsep}{0pt}\colorbox{small}{$2,400$}}  \\
\addlinespace[0.5em]

    \textbf{kin} & $53,955$  & {\setlength{\fboxsep}{0pt}\colorbox{srcsmall}{$3,122$}}&-&$70,307$& {\setlength{\fboxsep}{0pt}\colorbox{srclarge}{$85,824$}}&$83,898$&$77,271$&$62,236$\\
    
    \addlinespace[0.5em]

    \textbf{xho} & $60,557$ & {\setlength{\fboxsep}{0pt}\colorbox{srcsmall}{$3,439$}}&$70,506$&-&$64,179$& {\setlength{\fboxsep}{0pt}\colorbox{srclarge}{$125,604$}}&$138,111$&$113,453$\\
    
    \addlinespace[0.5em]

  \textbf{yor} & {\setlength{\fboxsep}{0pt}\colorbox{srclarge}{$133,353$}}& {\setlength{\fboxsep}{0pt}\colorbox{srcsmall}{$5,485$}}&$83,866$  &$62,471$  &-&$117,875$&$122,554$&$97,000$\\
    
    \addlinespace[0.5em]

    \textbf{swa} &$69,633$& {\setlength{\fboxsep}{0pt}\colorbox{srcsmall}{$3,507$}}&$84,025$&$125,307$&$121,233$&-& {\setlength{\fboxsep}{0pt}\colorbox{large}{$186,622$}}&$128,428$\\
 \addlinespace[0.5em]

   \textbf{eng} &$87,716$& {\setlength{\fboxsep}{0pt}\colorbox{srcsmall}{$5,692$}}&$77,148$&$137,240$&$125,927$& {\setlength{\fboxsep}{0pt}\colorbox{srclarge}{$186,122$}}&-&-\\
   \addlinespace[0.5em]

   \textbf{fra} &$58,521$& {\setlength{\fboxsep}{0pt}\colorbox{srcsmall}{$2,444$}}&$61,986$&$112,549$&$98,986$& {\setlength{\fboxsep}{0pt}\colorbox{srclarge}{$127,718$}}&-&-\\
    \bottomrule
  \end{tabular}
  }
  \caption{Number of parallel samples for each language direction. We highlight the {\setlength{\fboxsep}{0pt}\colorbox{large}{largest}} and {\setlength{\fboxsep}{0pt}\colorbox{small}{smallest}} parallel samples. We see for example that much more research on machine translation and data collation has been carried out on swa$\longleftrightarrow$eng than fon$\longleftrightarrow$fra, attesting to the under-representation of some African languages. }
  % Language codes are ISO 639-3.
  \label{tab:languagesize}
  \end{center}
\end{table*}
We devised preprocessing guidelines for each of our focus languages taking their linguistic properties into consideration. We used a maximum sequence length of $50$ (due to computational resources) and a minimum of $2$. In the following sections we will describe the data sources for the the parallel and monolingual corpora. 
\paragraph{Parallel Corpora:}
As NMT models are very reliant on parallel data, we sought to gather more parallel sentences for each language direction in an effort to increase the size and domain of each language direction. To this end, our first source was JW300~\cite{jw300}, a parallel corpus of over 300 languages with around 100 thousand biblical domain parallel sentences per language pair on average. Using OpusTools~\cite{opustools} we were able to get only very trustworthy translations by setting $t=1.5$ ($t$ is a threshold which indicates the confidence of the translations). We collected more parallel sentences from Tatoeba\footnote{\url{https://opus.nlpl.eu/Tatoeba.php}}, kde4\footnote{\url{https://huggingface.co/datasets/kde4}}~\cite{opus}, and some English-based bilingual samples from MultiParaCrawl\footnote{\url{https://www.paracrawl.eu/}}. 

Finally, following pointers from the native speakers of these focus languages in the Masakhane community~\cite{masakhane} to existing research on machine translation for African languages which open-sourced their parallel data, we assembled more parallel sentences mostly in the $\{en,fr\}$$\longleftrightarrow$AL direction.

From all this we created \code{MMTAfrica Test Set} (explained in more details in section~\ref{datasettype}), got $5,424,578$ total training samples for all languages directions (a breakdown of data size for each language direction is provided in Table~\ref{tab:languagesize}) and $4,000$ for dev.
\raggedbottom
\paragraph{Monolingual Corpora:} Despite our efforts to gather several parallel data from various domains, we were faced with some problems: 1) there was a huge imbalance in parallel samples across the language directions. In Table~\ref{tab:languagesize} we see that the $\circledast\longleftrightarrow$fon direction has the least amount of parallel sentences while $\circledast\longleftrightarrow$swa or $\circledast\longleftrightarrow$yor is made up of relatively larger parallel sentences. 2) the parallel sentences particularly for AL$\longleftrightarrow$AL span a very small domain (mostly biblical, internet )

We therefore set out to gather monolingual data from diverse sources. As our focus is on African languages, we collated monolingual data in only these languages. 

The monolingual sources and volume are summarized in Table~\ref{monosource}.
\begin{table}[H]
 \begin{center}
 \resizebox{\columnwidth}{!}{
 %\scalebox{0.50}{
 \footnotesize
  \begin{tabular}{p{2cm}p{5cm}p{1cm}}
    \toprule
    \textbf{Language(ID)} &\textbf{Monolingual source}&\textbf{Size}\\
    \midrule
   Xhosa (xho) &The CC100-Xhosa Dataset created by \citet{ccxhosa}, and OpenSLR~\cite{xhosa}&$158,660$ \\
    \addlinespace[0.5em]
    Yoruba (yor)& Yoruba Embeddings Corpus \cite{yoruba1} and MENYO20k \cite{yoruba2}&  $45,218$ \\
    \addlinespace[0.5em]
    Fon/Fongbe (fon) & FFR Dataset \cite{ffr}, and Fon French Daily Dialogues Parallel Data \cite{ffr_daily}& $42,057$\\
   \addlinespace[0.5em]
   Swahili/Kiswahili (swa) & \cite{swahili}& $23,170$ \\
   \addlinespace[0.5em]
  Kinyarwanda (kin)&KINNEWS-and-KIRNEWS \cite{kinyarwanda}&$7,586$ \\
    \addlinespace[0.5em]
  Igbo (ibo) & \cite{DBLP:journals/corr/abs-2004-00648}& $7,817$\\
    \bottomrule
  \end{tabular}
  }
  \caption{Monolingual data sources and sizes (number of samples).}
  % Language codes are ISO 639-3.
  \label{monosource}
  \end{center}
\end{table}

\subsection{Data Set Types in our Work}
\label{datasettype}
Here we elaborate on the different categories of data set that we (generated and) used in our work for training and evaluation. 
\begin{itemize}
\item \code{FLORES Test Set}: This refers to the dev test set of $1012$ parallel sentences in all $101$ language directions provided by ~\citet{flores101}\footnote{\url{https://dl.fbaipublicfiles.com/flores101/dataset/flores101_dataset.tar.gz}}. We performed evaluation on this test set for all language directions except $\circledast\longleftrightarrow$fon and $\circledast\longleftrightarrow$kin.

\item \code{MMTAfrica Test Set}: This is a test set we created by taking out a small but equal number of sentences from each parallel source domain.  As a result, we have a set from a wide range of domains, while encompassing samples from many existing test sets from previous research. Although this set is small to be fully considered as a test set, we open-source it because it contains sentences from many domains (making it useful for evaluation) and we hope that it can be built upon, by perhaps merging it with other benchmark test sets~\cite{abate-etal-2018-parallel,abbott-martinus-2019-benchmarking,reid2021afromt}. 

\item \code{Baseline Train/Test Set}: We first conducted baseline experiments with Fon, Igbo, English and French as explained in section~\ref{baseline}. For this we created a special data set by carefully selecting a small subset of the FFR Dataset (which already contained parallel sentences in French and Fon), first automatically translating the sentences to English and Igbo, using the Google Translate API\footnote{\url{https://cloud.google.com/translate}}, and finally re-translating with the help of Igbo (7) and English (7) native speakers (we recognized that it was easier for native speakers to edit/tweak an existing translation rather than writing the whole translation from scratch). In so doing, we created a data set of $13,878$ translations in all $4$ language directions. We split the data set into $12,554$ for training \code{Baseline Train Set}, $662$ for dev and $662$ for test \code{Baseline Test Set}.
\end{itemize}
\section{Model and Training Setup}
\label{trainingobj}
For each language direction $X\rightarrow Y$ we have its set of $n$ parallel sentences $\mathcal{D}=\{(x_{i},y_{i})\}_{i=1}^{n}$ where $x_{i}$ is the $i$th source sentence of language $X$ and $y_{i}$ is its translation in the target language $Y$.

Following the approach of~\citet{johnson-etal-2017-googles} and \citet{mt5}, we model translation in a text-to-text format. More specifically, we create the input for the model by prepending the target language tag to  the source sentence. Therefore for each source sentence $x_{i}$ the input to the model is \code{<$Y_{tag}$> $x_{i}$} and the target is $y_{i}$. Taking a real example, let's say we wish to translate the Igbo sentence \textit{Daal\d{u} maka ikwu eziokwu nke Chineke} to English. The input to the model becomes \textit{<eng> Daal\d{u} maka ikwu eziokwu nke Chineke}. 

\subsection{Model Setup}
For all our experiments, we used the mT5 model~\cite{mt5}, a multilingual  variant of the encoder-decoder, transformer-based~\cite{transformer} “Text-to-Text Transfer Transformer” (T5) model~\cite{t5}. In T5 pre-training, the NLP tasks (including machine translation) were cast into a “text-to-text” format -- that is, a task where the model is fed some text prefix for context or conditioning and is then asked to produce some output text. This framework makes it straightforward to design a number of NLP tasks like machine translation, summarization, text classification, etc. Also, it provides a consistent training objective both for pre-training and finetuning. The mT5 model was pre-trained with a maximum likelihood objective using “teacher forcing”~\cite{teacherforcing}. The mT5 model was also pretrained with a modification of the masked language modelling objective~\cite{bert}.

We finetuned the \code{mt5-base} model on our many-to-many machine translation task.  While \citet{mt5} suggest that higher versions of the mT5 model (\textit{Large}, \textit{XL} or \textit{XXL}) give better performance on downstream multilingual translation tasks, we  were constrained by computational resources to \code{mt5-base}, which has $580M$ parameters.

\subsection{Training Setup}
We have a set of language tags $L$ for the languages we are working with in our multilingual many-to-many translation. In our baseline setup (section~\ref{baseline}) $L=\{eng,fra,ibo,fon\}$ and in our final experiment (section~\ref{all}) $L=\{eng,fra,ibo,fon,swa,kin,xho,yor\}$. We carried out many-to-many translation using all the possible directions from $L$ except $eng\longleftrightarrow fra$. We skipped $eng\longleftrightarrow fra$ for this fundamental reason: 
\begin{itemize}
    \item our main focus is on African$\longleftrightarrow$African or $\{eng,fra\}$ $\longleftrightarrow$African. Due to the high-resource nature of English and French, adding the training set for $eng\longleftrightarrow fra$ would overshadow the learning of the other language directions and greatly impede our analyses. Our intuition draws from the observation of \citet{mt5} as the reason for \textit{off-target} translation in the mT5 model: as English-based finetuning proceeds, the model’s assigned likelihood of non-English tokens presumably decreases. Therefore since the \code{mt5-base} training set contained predominantly English (and after other European languages) tokens and our research is about AL$\longleftrightarrow$AL translation, removing the $eng\longleftrightarrow fra$ direction was our way of ensuring the model designated more likelihood to AL tokens.
\end{itemize}

\subsubsection{Our Contributions}
\par

In addition to the parallel data between the African languages, we leveraged monolingual data to improve translation quality in two ways:
\begin{enumerate}
    \item \textbf{our backtranslation (BT):}  We designed a modified form of the random online backtranslation~\cite{DBLP:journals/corr/abs-2004-11867} where instead of randomly selecting a subset of languages to backtranslate, we selected for each language $num\_bt$ sentences at random from the monolingual data set. This means that the model gets to backtranslate different (monolingual) sentences every backtranslation time and in so doing, we believe, improve the model's domain adaptation because it gets to learn from various samples from the whole monolingual data set. We initially tested different values of $num\_bt$ to find a compromise between backtranslation computation time and translation quality. Following research works which have shown the effectiveness of random beam-search over greedy decoding while generating backtranslations~\cite{lampe,understandingbt,hoang-etal-2018-iterative,DBLP:journals/corr/abs-2004-11867}, we generated $num\_sample$ prediction sentences from the model and randomly selected (with equal probability) one for our backtranslated sentence. Naturally the value of $num\_sample$ further affects the computation time (because the model has to produce $num\_sample$ different output sentences for each input sentence) and so we finally settled with $num\_sample=2$.
    \item \textbf{our reconstruction:} Given a monolingual sentence $x^{m}$ from language $m$, we applied random swapping ($2$ times) and deletion (with a probability of $0.2$) to get a noisy version $\hat{x}$. Taking inspiration from \citet{t5} we integrated the reconstruction objective into our model finetuning by prepending the language tag <$m$> to $\hat{x}$ and setting its target output to $x^{m}$. 
\end{enumerate}

\section {Experiments}
In all our experiments we initialized the pretrained \code{mT5-base} model using Hugging Face's AutoModelForSeq2SeqLM and tracked the training process with Weights\&Biases~\cite{wandb}. We used the AdamW optimizer~\cite{DBLP:journals/corr/abs-1711-05101} with a learning rate (lr) of $3e^{-6}$ and transformer's $get\_linear\_schedule\_with\_warmup$ scheduler (where the learning rate decreases linearly from the initial lr set in the optimizer to $0$, after a warmup period and then increases linearly from $0$ to the initial lr set in the optimizer.)
\subsubsection{Baseline}
\label{baseline}
The goal of our baseline was to understand the effect of jointly finetuning with backtranslation and reconstruction on the African$\longleftrightarrow$African language translation quality in two scenarios: when the AL was initially pretrained on the multilingual model and contrariwise. Using Fon (which was not initially included in the pretraining) and Igbo (which was initially included in the pretraining) as the African languages for our baseline training, we finetuned our model on a many-to-many translation in all directions of $\{eng,fra,ibo,fon\}/ eng\longleftrightarrow fra$ amounting to $10$ directions. We used the \code{Baseline Train Set} for training and the \code{Baseline Test Set} for evaluation. We trained the model for only 3 epochs in three settings:
\begin{enumerate}
    \item \code{BASE}: in this setup we finetune the model on only the many-to-many translation task: no backtranslation nor reconstruction.
    \item \label{btonly} \code{BT}: refers to finetuning with our backtranslation objective described in section~\ref{trainingobj}. For our baseline, where we backtranslate using monolingual data in $\{ibo,fon\}$, we set $num\_bt=500$. For our final experiments, we first tried with $500$ but finally reduced to $100$ due to the great deal of computation required. For our baseline experiment, we ran one epoch normally and the remaining two with backtranslation. For our final experiments, we first finetuned the model on $3$ epochs before continuing with backtranslation.
    \item  \code{BT\&REC}: refers to joint backtranslation and reconstruction (explained in section~\ref{trainingobj}) while finetuning. Two important questions were addressed -- 1) the ratio, backtranslation : reconstruction, of monolingual sentences to use  and 2) whether to use the same or different sentences for backtranslation and reconstruction. Bearing computation time in mind, we resolved to go with $500:50$ for our baseline and $100:50$ for our final experiments. We leave ablation studies on the effect of the ratio on translation quality to future work. For the second question we decided to randomly sample (with replacement) different sentences each for our backtranslation and reconstruction.
\end{enumerate}

For our baseline, we used a learning rate of $5e^{-4}$, a batch size of 32 sentences, with gradient accumulation up to a batch of 256 sentences and an early stopping patience of 100 evaluation steps. To further analyse the performance of our baseline setups we ran \code{comparemt}\footnote{\url{https://github.com/neulab/compare-mt}} \cite{comparemt} on the model's predictions.

\subsubsection{MMTAfrica}
\label{all}
MMTAfrica refers to our final experimental setup where we finetuned our model on all language directions involving all eight languages  $L=\{eng,fra,ibo,fon,swa,kin,xho,yor\}$ except eng$\longleftrightarrow$fra. Taking inspiration from our baseline results we ran our experiment with our proposed \code{BT\&REC} setting and made some adjustments along the way.

The long computation time for backtranslating (with just 100 sentences per language the model was required to generate around $3,000$ translations every backtranslation time) was a drawback. To mitigate the issue we parallelized the process using the multiprocessing package in Python\footnote{\url{https://docs.python.org/3/library/multiprocessing.html}}. We further slowly reduced the number of sentences for backtranslation (to $50$, and finally $10$).

Gradient descent in large multilingual models has been shown to be more stable when updates are performed over large batch sizes are used~\cite{mt5}. To cope with our computational resources, we used gradient accumulation to increase updates from an initial batch size of $64$ sentences, up to a batch gradient computation size of $4096$ sentences. We further utilized PyTorch's DataParallel package\footnote{\url{https://pytorch.org/docs/stable/generated/torch.nn.DataParallel.html}} to parallelize the training across the GPUs. We used a learning rate (lr) of $3e^{-6}$

\section{Results and Insights}
All evaluations were made using spBLEU (sentencepiece~\cite{sentencepiece} + sacreBLEU~\cite{sacrebleu}) as described in \cite{flores101}. We further evaluated on the chrF~\cite{chrf} and TER metrics.
\subsection{Baseline Results and Insights}
\label{baselineresults}
\begin{figure}[h]
\includegraphics[width=\linewidth]{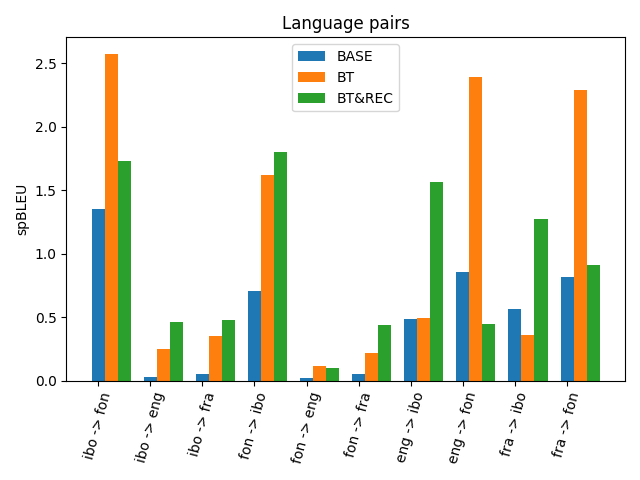}
\centering
\caption{\label{base-bt-rec} spBLEU scores of the 3 setups explained in section~\ref{baseline}}
\end{figure}

Figure~\ref{base-bt-rec} compares the spBLEU scores for the three setups used in our baseline experiments. As a reminder, we make use of the symbol $\diamond$ to refer to any language in the set $\{eng,fra,ibo,fon\}$.

\code{BT} gives strong improvement over \code{BASE} (except in eng$\longrightarrow$ibo where it's relatively the same, and fra$\longrightarrow$ibo where it performs worse).

When the target language is \textbf{\textit{fon}}, we observe a considerable boost in the spBLEU of the \code{BT} setting, which also significantly outperformed \code{BASE} and \code{BT\&REC}. \code{BT\&REC} contributed very little when compared with \code{BT} and sometimes even performed poorly (in eng$\longrightarrow$fon). We attribute this poor performance from the reconstruction objective to the fact that the \code{mt5-base} model was not originally pretrained on Fon. Therefore, with only 3 epochs of finetuning (and 1 epoch before introducing the reconstruction and backtranslation objectives) the model was not able to meaningfully utilize both objectives.

Conversely, when the target language is \textbf{\textit{ibo}} \code{BT\&REC} gives best results -- even in scenarios where \code{BT} underperforms \code{BASE}(as is the case of fra$\longrightarrow$ibo and eng$\longrightarrow$ibo). We believe that the decoder of the model, being originally pretrained on corpora containing Igbo, was able to better use our reconstruction to improve translation quaity in $\diamond\longrightarrow$\textbf{\textit{ibo}} direction. 

Drawing insights from fon$\longleftrightarrow$ibo \textbf{we offer the following propositions concerning AL$\longleftrightarrow$AL multilingual translation}:
\begin{itemize}
    \item our backtranslation (section~\ref{trainingobj}) from monolingual data improves the cross-lingual mapping of the model for low-resource African languages. While it is computationally expensive, our parallelization and decay of number of backtranslated sentences are some potential solutions towards effectively adopting backtranslation using monolingual data. 
\item Denoising objectives typically have been known to improve machine translation quality~\cite{zhang-zong-2016-exploiting,cheng-etal-2016-semi,gu,DBLP:journals/corr/abs-2004-11867,mt5} because they imbue the model with more generalizable knowledge (about that language) which is used by the decoder to predict better token likelihoods for that language during translation. This is a reasonable explanation for the improved quality with the \code{BT\&REC} over \code{BT} in the $\diamond\longrightarrow$\textbf{\textit{ibo}}. As we learned from $\diamond\longrightarrow$\textbf{\textit{fon}}, using reconstruction could perform unsatisfactorily if not handled well. Some methods we propose are:
\begin{enumerate}
    \item For African languages that were included in the original model pretraining (as was the case of Igbo, Swahili, Xhosa, and \yoruba in the mT5 model), using the \code{BT\&REC} setting for finetuning produces best results. While we did not perform ablation studies on the data size ratio for backtranslation and reconstruction, we believe that our ratio of $2:1$ (in our final experiments) gives the best compromise on both computation time and translation quality.
    \item For African languages that were not originally included in the original model pretraining (as was the case of Kinyarwanda and Fon in the mT5 model), reconstruction together with backtranslation (especially at an early stage) only introduces more noise which could harm the cross-lingual learning. For these languages we propose: 
    \begin{enumerate}
        \item first finetuning the model on only our reconstruction (described in section~\ref{trainingobj}) for fairly long training steps before using \code{BT\&REC}. This way, the initial reconstruction will help the model learn that language representation space and increase its the likelihood of tokens.
          \end{enumerate}
\end{enumerate}
\end{itemize}
\subsection{MMTAfrica Results and Insights}
\label{mmtafricaresult}
In Table~\ref{floresTest}, we compared MMTAfrica with the M2M MMT\cite{m2m} benchmark results of~\citet{flores101} using the same test set they used -- \code{FLORES Test Set}. On all language pairs except swa$\rightarrow$eng (which has a comparable $-2.76$ spBLEU difference), we report an improvement from MMTAfrica (spBLEU gains ranging from $+0.58$ in swa$\longrightarrow$fra to $+19.46$ in fra$\longrightarrow$xho). The lower score of swa$\rightarrow$eng presents an intriguing anomaly, especially given the large availability of parallel corpora in our training set for this pair. We plan to investigate this in further work.
\begin{table}[ht]
\begin{center}
 \resizebox{\columnwidth}{!}{
  \footnotesize
  \begin{tabular}{ccccc}
    \toprule
    \textbf{Source} &\textbf{Target}&\textcolor{purple}{\textbf{spBLEU (FLORES)}}$\uparrow$ & \textcolor{purple}{\textbf{spBLEU (Ours*)}}$\uparrow$ & \textbf{spCHRF (Ours*)}$\uparrow$\\
    \midrule
ibo&swa&\textcolor{black}{4.38}&\textcolor{black}{\textbf{21.84}} | 11.63&37.38 | 35.66\\
\addlinespace[0.5em]
ibo&xho&\textcolor{black}{2.44}&\textcolor{black}{\textbf{13.97}} | 7.65&31.95 | 29.47\\
\addlinespace[0.5em]
ibo&yor&\textcolor{black}{1.54}&\textcolor{black}{\textbf{10.72}} | 7.72&26.55 | 16.84\\
\addlinespace[0.5em]
ibo&eng&\textcolor{black}{7.37}&\textcolor{black}{\textbf{13.62}} | 15.44&38.90 | 37.99\\
\addlinespace[0.5em]
ibo&fra&\textcolor{black}{6.02}&\textcolor{black}{\textbf{16.46}} | 12.89&35.10 | 31.71\\
\addlinespace[0.5em]
\hline
\addlinespace[0.5em]
swa&ibo&\textcolor{black}{1.97}&\textcolor{black}{\textbf{19.80}} | 16.73&33.95 | 28.07\\
\addlinespace[0.5em]
swa&xho&\textcolor{black}{2.71}&\textcolor{black}{\textbf{21.71}} | 11.74&39.86 | 35.67\\
\addlinespace[0.5em]
swa&yor&\textcolor{black}{1.29}&\textcolor{black}{\textbf{11.68}} | 8.28&27.44 | 17.18\\
\addlinespace[0.5em]
swa&eng&\textcolor{black}{\textbf{30.43}}&\textcolor{black}{27.67} | 28.41&56.12 | 53.65\\
\addlinespace[0.5em]
swa&fra&\textcolor{black}{26.69}&\textcolor{black}{\textbf{27.27}} | 19.85&46.20 | 41.41\\
\addlinespace[0.5em]
\hline
\addlinespace[0.5em]
xho&ibo&\textcolor{black}{3.80}&\textcolor{black}{\textbf{17.02}} | 15.28&31.30 | 26.67\\
\addlinespace[0.5em]
xho&swa&\textcolor{black}{6.14}&\textcolor{black}{\textbf{29.47}} | 15.73&44.68 | 40.78\\
\addlinespace[0.5em]
xho&yor&\textcolor{black}{1.92}&\textcolor{black}{\textbf{10.42}} | 7.82&26.77 | 17.10\\
\addlinespace[0.5em]
xho&eng&\textcolor{black}{10.86}&\textcolor{black}{\textbf{20.77}} | 21.75&48.69 | 46.34\\
\addlinespace[0.5em]
xho&fra&\textcolor{black}{8.28}&\textcolor{black}{\textbf{21.48}} | 15.97&40.65 | 36.28\\
\addlinespace[0.5em]
\hline
\addlinespace[0.5em]
yor&ibo&\textcolor{black}{1.85}&\textcolor{black}{\textbf{11.45}} | 11.44&25.26 | 21.70\\
\addlinespace[0.5em]
yor&swa&\textcolor{black}{1.93}&\textcolor{black}{\textbf{14.99}} | 6.61&30.49 | 28.21\\
\addlinespace[0.5em]
yor&xho&\textcolor{black}{1.94}&\textcolor{black}{\textbf{9.31}} | 4.99&26.34 | 24.27\\
\addlinespace[0.5em]
yor&eng&\textcolor{black}{4.18}&\textcolor{black}{\textbf{8.15}} | 9.02&30.65 | 28.85\\
\addlinespace[0.5em]
yor&fra&\textcolor{black}{3.57}&\textcolor{black}{\textbf{10.59}} | 7.91&27.60 | 23.93\\
\addlinespace[0.5em]
\hline
\addlinespace[0.5em]
eng&ibo&\textcolor{black}{3.53}&\textcolor{black}{\textbf{21.49}} | 19.52&37.24 | 32.46\\
\addlinespace[0.5em]
eng&swa&\textcolor{black}{26.95}&\textcolor{black}{\textbf{40.11}} | 27.06&53.13 | 51.90\\
\addlinespace[0.5em]
eng&xho&\textcolor{black}{4.47}&\textcolor{black}{\textbf{27.15}} | 14.85&44.93 | 39.88\\
\addlinespace[0.5em]
eng&yor&\textcolor{black}{2.17}&\textcolor{black}{\textbf{12.09}} | 9.43&28.34 | 18.39\\
\addlinespace[0.5em]
\hline
\addlinespace[0.5em]
fra&ibo&\textcolor{black}{1.69}&\textcolor{black}{\textbf{19.48}} | 17.25&34.47 | 29.49\\
\addlinespace[0.5em]
fra&swa&\textcolor{black}{17.17}&\textcolor{black}{\textbf{34.21}} | 19.49&48.95 | 45.44\\
\addlinespace[0.5em]
fra&xho&\textcolor{black}{2.27}&\textcolor{black}{\textbf{21.73}} | 11.37&40.06 | 35.41\\
\addlinespace[0.5em]
fra&yor&\textcolor{black}{1.16}&\textcolor{black}{\textbf{11.42}} | 8.54&27.67 | 17.53\\
\bottomrule
  \end{tabular}
  }
    \end{center}

  \caption{Evaluation Scores of the Flores M2M MMT model and MMTAfrica on \code{FLORES Test Set}. We use \code{|} to denote spBLEU with | without BT\&REC.}
  \label{floresTest}
\end{table}

In Table~\ref{OurTest} we introduce benchmark results of MMTAfrica on \code{MMTAfrica Test Set} with and without BT\&REC. We also put the test size of each language pair. The spBLEU scores demonstrate the efficiency of our new objective, as it led to improvements in majority of the tasks.
\paragraph{Interesting analysis about Fon (fon) and \yoruba (yor):}
For each language, the lowest spBLEU scores in both tables come from the $\longrightarrow$yor direction, except fon$\longleftrightarrow$yor (from Table~\ref{OurTest}) which interestingly has the highest spBLEU score compared to the other fon$\longrightarrow\circledast$ directions. We do not know the reason for the very low performance in the $\circledast\longrightarrow$yor direction, but we offer below a plausible explanation about fon$\longleftrightarrow$yor.

The oral linguistic history of Fon ties it to the ancient \yoruba kingdom~\cite{barnes1997africa}. Furthermore, in present day Benin, where Fon is largely spoken as a native language, Yoruba is one of the indigenuous languages commonly spoken. \footnote{\url{https://en.wikipedia.org/wiki/Benin} (Last Accessed : 30.08.2021).} Therefore Fon and \yoruba share some linguistic characteristics and we believe this is one logic behind the fon$\longleftrightarrow$yor surpassing other fon$\longrightarrow\circledast$ directions. 

This explanation could inspire transfer learning from \yoruba, which has received comparably more research and has more resources for machine translation, to Fon. We leave this for future work.
\raggedbottom
\section{Conclusion and Future Work}
In this paper, we introduced MMTAfrica, a multilingual machine translation model on 6 African Languages. Our results and analyses, including a new reconstruction objective, give insights on MMT for African languages for future research.

\begin{table}[H]
 \begin{center}
 \resizebox{\columnwidth}{!}{
 %\scalebox{0.90}{
 \footnotesize
  \begin{tabular}{cccccc}
    \toprule
    \textbf{Source} &\textbf{Target}&\textbf{Test size}&  \textbf{spBLEU}$\uparrow$ & \textbf{spCHRF}$\uparrow$ & \textbf{spTER}$\downarrow$\\
    \midrule
ibo&swa&60&34.89 (12.27)&47.38 (36.65)&68.28 (124.01)\\
\addlinespace[0.5em]
ibo&xho&30&36.69 (21.92)&50.66 (41.40)&59.65 (76.36)\\
\addlinespace[0.5em]
ibo&yor&30&11.77 (10.19)&29.54 (22.10)&129.84 (130.39)\\
\addlinespace[0.5em]
ibo&kin&30&33.92 (16.07)&46.53 (36.95)&67.73 (96.5)\\
\addlinespace[0.5em]
ibo&fon&30&35.96 (11.47)&43.14 (21.75)&63.21 (91.91)\\
\addlinespace[0.5em]
ibo&eng&90&37.28 (11.70)&60.42 (38.11)&62.05 (110.67)\\
\addlinespace[0.5em]
ibo&fra&60&30.86 (6.02)&44.09 (28.13)&69.53 (121.43)\\
\addlinespace[0.5em]
\hline
\addlinespace[0.5em]
swa&ibo&60&33.71 (23.12)&43.02 (33.91)&60.01 (85.18)\\
\addlinespace[0.5em]
swa&xho&30&37.28 (20.55)&52.53 (40.84)&55.86 (72.71)\\
\addlinespace[0.5em]
swa&yor&30&14.09 (15.49)&27.50 (23.50)&113.63 (106.22)\\
\addlinespace[0.5em]
swa&kin&30&23.86 (13.53)&42.59 (36.88)&94.67 (118.0)\\
\addlinespace[0.5em]
swa&fon&30&23.29 (8.94)&33.52 (16.97)&65.11 (84.12)\\
\addlinespace[0.5em]
swa&eng&60&35.55 (43.11)&60.47 (66.52)&47.32 (40.0)\\
\addlinespace[0.5em]
swa&fra&60&30.11 (21.99)&48.33 (43.84)&63.38 (71.17)\\
\addlinespace[0.5em]
\hline
\addlinespace[0.5em]
xho&ibo&30&33.25 (24.33)&45.36 (36.42)&62.83 (70.63)\\
\addlinespace[0.5em]
xho&swa&30&39.26 (23.42)&53.75 (46.22)&53.72 (67.13)\\
\addlinespace[0.5em]
xho&yor&30&22.00 (16.86)&38.06 (27.20)&70.45 (74.36)\\
\addlinespace[0.5em]
xho&kin&30&30.66 (14.09)&46.19 (37.26)&74.70 (112.4)\\
\addlinespace[0.5em]
xho&fon&30&25.80 (10.73)&34.87 (18.70)&65.96 (85.51)\\
\addlinespace[0.5em]
xho&eng&90&30.25 (21.36)&55.12 (48.96)&62.11 (69.61)\\
\addlinespace[0.5em]
xho&fra&30&29.45 (16.25)&45.72 (35.99)&61.03 (70.91)\\
\addlinespace[0.5em]
\hline
\addlinespace[0.5em]
yor&ibo&30&25.11 (15.00)&34.19 (26.75)&74.80 (97.44)\\
\addlinespace[0.5em]
yor&swa&30&17.62 (4.81)&34.71 (28.23)&85.18 (130.22)\\
\addlinespace[0.5em]
yor&xho&30&29.31 (14.43)&43.13 (32.34)&66.82 (87.2)\\
\addlinespace[0.5em]
yor&kin&30&25.16 (14.65)&38.02 (32.22)&72.67 (86.99)\\
\addlinespace[0.5em]
yor&fon&30&31.81 (10.28)&37.45 (17.52)&63.39 (88.57)\\
\addlinespace[0.5em]
yor&eng&90&17.81 (2.11)&41.73 (22.90)&93.00 (93.49)\\
\addlinespace[0.5em]
yor&fra&30&15.44 (5.62)&30.97 (22.81)&90.57 (136.96)\\
\addlinespace[0.5em]
\hline
\addlinespace[0.5em]
kin&ibo&30&31.25 (27.30)&42.36 (37.44)&66.73 (76.1)\\
\addlinespace[0.5em]
kin&swa&30&33.65 (13.00)&46.34 (38.62)&72.70 (100.84)\\
\addlinespace[0.5em]
kin&xho&30&20.40 (9.96)&39.71 (33.27)&89.97 (108.05)\\
\addlinespace[0.5em]
kin&yor&30&18.34 (17.64)&33.53 (27.48)&70.43 (72.4)\\
\addlinespace[0.5em]
kin&fon&30&22.43 (10.84)&32.49 (18.40)&67.26 (82.65)\\
\addlinespace[0.5em]
kin&eng&60&15.82 (9.28)&43.10 (35.88)&96.55 (102.82)\\
\addlinespace[0.5em]
kin&fra&30&16.23 (12.24)&33.51 (29.41)&91.82 (100.75)\\
\addlinespace[0.5em]
\hline
\addlinespace[0.5em]
fon&ibo&30&32.36 (16.24)&46.44 (31.18)&61.82 (83.54)\\
\addlinespace[0.5em]
fon&swa&30&29.84 (17.08)&42.96 (35.26)&72.28 (88.37)\\
\addlinespace[0.5em]
fon&xho&30&28.82 (13.59)&43.74 (31.80)&66.98 (93.27)\\
\addlinespace[0.5em]
fon&yor&30&30.45 (22.17)&42.63 (30.52)&60.72 (70.91)\\
\addlinespace[0.5em]
fon&kin&30&23.88 (10.08)&39.59 (28.96)&78.06 (91.81)\\
\addlinespace[0.5em]
fon&eng&30&16.63 (13.67)&41.63 (30.36)&69.03 (83.57)\\
\addlinespace[0.5em]
fon&fra&60&24.79 (17.31)&43.39 (33.39)&82.15 (82.97)\\
\addlinespace[0.5em]
\hline
\addlinespace[0.5em]
eng&ibo&90&44.24 (25.18)&54.89 (35.84)&63.92 (87.47)\\
\addlinespace[0.5em]
eng&swa&60&49.94 (33.53)&61.45 (55.58)&47.83 (65.22)\\
\addlinespace[0.5em]
eng&xho&120&31.97 (22.57)&49.74 (46.01)&72.89 (68.47)\\
\addlinespace[0.5em]
eng&yor&90&23.93 (11.01)&36.19 (19.25)&84.05 (90.29)\\
\addlinespace[0.5em]
eng&kin&90&40.98 (11.47)&56.00 (30.05)&76.37 (101.42)\\
\addlinespace[0.5em]
eng&fon&30&27.19 (6.40)&36.86 (14.91)&62.54 (91.08)\\
\addlinespace[0.5em]
\hline
\addlinespace[0.5em]
fra&ibo&60&36.47 (18.26)&46.93 (28.72)&59.91 (86.05)\\
\addlinespace[0.5em]
fra&swa&60&36.53 (20.72)&51.42 (46.35)&55.94 (66.86)\\
\addlinespace[0.5em]
fra&xho&30&34.35 (21.49)&49.39 (43.36)&60.30 (72.39)\\
\addlinespace[0.5em]
fra&yor&30&7.26 (7.88)&25.54 (19.59)&124.53 (121.17)\\
\addlinespace[0.5em]
fra&kin&30&31.07 (17.24)&42.26 (37.63)&81.06 (95.45)\\
\addlinespace[0.5em]
fra&fon&60&31.07 (10.82)&38.72 (21.10)&75.74 (93.33)\\
\addlinespace[0.5em]
    \bottomrule
  \end{tabular}
  }
  \caption{Benchmark Evaluation Scores on \code{MMTAfrica Test Set} with (without) BT\&REC}
  % Language codes are ISO 639-3.
  \label{OurTest}
  \end{center}
\end{table}
\newpage
\bibliography{anthology,custom}
\bibliographystyle{acl_natbib}

%f we have time, we should do some human evaluation.
%\section{Discussion and Future works}
% \appendix

% \section{Example Appendix}
% \label{sec:appendix}

% This is an appendix.

\end{document}